\documentclass[letterpaper]{article}
\usepackage[preprint]{aaai2027}
\nocopyright
\usepackage[hyphens]{url}
\usepackage{graphicx}
\usepackage{natbib}
\usepackage{caption}
\usepackage{float}
\usepackage{booktabs}
\usepackage{tabularx}
\usepackage{cuted}
\usepackage{amsmath}
\usepackage{amssymb}
\title{Radar4D-VLM: Proposal-Grounded Temporal 4D Radar Reasoning Across Frozen Language Models}
\author{Jiaju Han\textsuperscript{1,2},
Xuemeng Sun\textsuperscript{1},
Qike Zhang\textsuperscript{1},
Xiang Chen\textsuperscript{1},
Luwei Yang\textsuperscript{2},\\
Jiahuan Long\textsuperscript{3},
Yiwei Wei\textsuperscript{4},
Jiujiang Guo\textsuperscript{4,5},
Chengyin Hu\textsuperscript{1}}
\affiliations{\textsuperscript{1}China University of Petroleum-Beijing at Karamay\\
\textsuperscript{2}Shenzhen Research Institute of Big Data \quad
\textsuperscript{3}Shanghai Jiao Tong University\\
\textsuperscript{4}Tianjin University \quad
\textsuperscript{5}North University of China}

\begin{document}

\maketitle

\begin{abstract}
Vision--language models for autonomous driving primarily rely on cameras and LiDAR, leaving 4D radar largely unexplored as a standalone perceptual modality despite its robustness to adverse visibility and direct measurement of radial velocity. We introduce Radar4D-VLM, a radar-only temporal vision--language model that reasons from ten consecutive 4D-radar point-cloud sweeps without camera or LiDAR input. Radar4D-VLM extracts geometrically grounded object proposals and organizes radar evidence into a compact hierarchy of object, scene, and kinematic tokens. A parameter-efficient projector maps these tokens into frozen language backbones, while auditable prediction heads jointly model object count, spatial distribution, motion state, collision risk, semantic category, and radial velocity. Radar4D-VLM combines proposal-grounded temporal object tokenization, global scene context, and explicit kinematic tokens within a unified frozen-backbone interface. On sequence-isolated K-Radar development validation, its Top-64 proposal recall reaches 98.13\% at 4~m, exceeding fixed-lattice and uniform-random controls by 6.40 and 22.83 percentage points, respectively. We further evaluate 24 matched runs spanning eight frozen Qwen, Phi, Mistral, Llama, and Gemma backbones under an identical adaptation budget. The radar-token interface remains compatible across all five language-model families, while matched aligned, permuted, and no-language controls show sensor dependence but no stable direct-head gain from aligned language supervision. These results establish a reproducible foundation for radar-only multimodal scene and motion reasoning while separating interface compatibility from the benefit of language supervision.
\end{abstract}

\section{Introduction}

Autonomous driving requires perception that remains reliable when visibility deteriorates. Cameras and LiDAR provide rich appearance and geometry, but their measurements degrade in fog, rain, and snow; 4D millimeter-wave radar remains operational and directly measures radial velocity \citep{paek2022kradar,hamilton2026weather}. Its point clouds are nevertheless sparse, nonuniform, and difficult to connect to object-level semantics and motion reasoning. This gap motivates a radar-only interface that preserves physical evidence while making it accessible to language models.

Recent radar--language research has established several complementary settings. The Radar Spectra--Language Model aligns raw automotive radar spectra with a vision--language embedding space for retrieval and scene parsing \citep{pushkareva2024radarspectra}. Talk2Radar treats language as an input query for referring-expression grounding in 4D radar point clouds \citep{guan2025talk2radar}, while RLM learns scene-level radar representations from simulated radar--caption pairs and structured spatial supervision \citep{mishra2025rlm}. Hamilton and Heckman map K-Radar features through a frozen vision--language model for structured captioning \citep{hamilton2026weather}; RadarLLM instead studies temporal millimeter-wave point clouds for human-motion understanding \citep{lai2026radarllm}. These works demonstrate that radar and language can be connected, but they leave two coupled questions open: how to organize temporal automotive radar into object-, scene-, and motion-aware tokens, and whether successful language outputs reflect sensor use or an actual benefit from aligned language supervision.

The second question requires controlled attribution. A model may produce plausible answers because its interface is compatible with a frozen backbone, because it uses real radar evidence, or because aligned language supervision improves the shared radar pathway. These are distinct claims. Output capability establishes an interface, while sensitivity to radar interventions establishes sensor dependence; neither alone shows that the language objective improves the underlying representation \citep{standley2020tasks,geirhos2020shortcut}. We therefore compare aligned supervision with a fixed within-task verbalizer permutation and no language objective on identical windows, and evaluate their shared direct heads independently of answer-token agreement.

Radar4D-VLM makes its representation and attribution explicitly auditable. A train-split RTNH-compatible encoder proposes 64 centers from ten sweeps; temporal and Doppler evidence yields 64 object, four scene, and one kinematic token. These tokens feed shared prediction heads and a budget-matched projector connected to a frozen language backbone. Inference excludes camera, LiDAR, text, tracks, and ground-truth boxes, separating radar dependence from supervision effects.
Our contributions are threefold:
\begin{itemize}
\item To the best of our knowledge, based on searches of arXiv, IEEE Xplore, and OpenReview through July 29, 2026, Radar4D-VLM is the first automotive radar--language architecture to combine proposal-grounded temporal object tokenization, global scene context, and an explicit kinematic token within a unified frozen-backbone interface. It converts ten radar sweeps into 64 object tokens, four scene tokens, and one kinematic token without camera or LiDAR input at inference.
\item We conduct an extensive controlled evaluation comprising a 24-run cross-family matrix over eight frozen backbones from five language-model families, paired aligned/permuted/no-language objectives, and radar-input interventions. The results show 98.13\% Top-64 proposal recall, compatibility across all audited backbones, and clear dependence on real radar and temporal order, but no stable direct-head gain from aligned language supervision.
\item We perform systematic ablations of temporal history, Doppler, proposal and scene features, global token pooling, and the kinematic token. They reveal that ten-sweep history and real radar content provide the strongest consistent signal, whereas Doppler and isolated kinematic-token effects are smaller or mixed. Exact tensor pruning further enables LLM-free direct inference within a registered GPU-nondeterminism envelope.
\end{itemize}

\begin{figure*}[t]
\centering
\includegraphics[width=\textwidth]{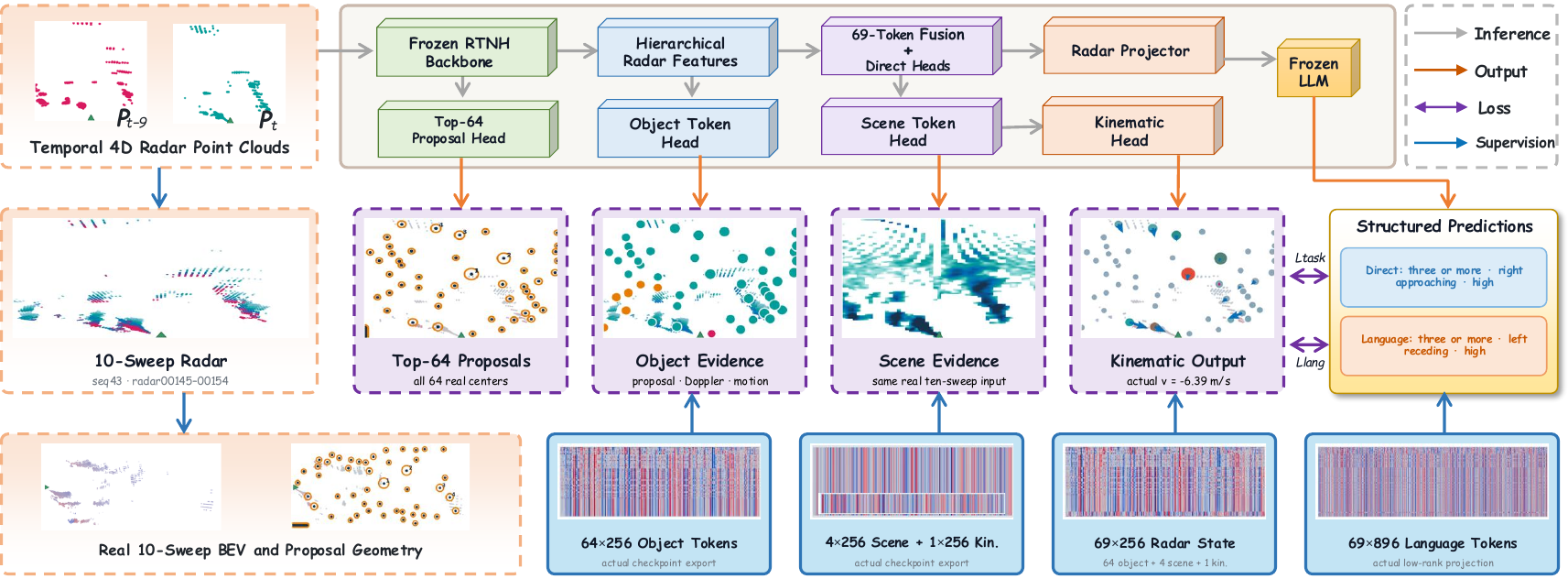}
\caption{Radar4D-VLM architecture and controlled training conditions. The current sweep is processed by a frozen, train-split RTNH-compatible proposal encoder, while all ten radar sweeps provide proposal-aligned temporal and Doppler evidence. Trainable modules construct 64 object tokens, four scene tokens, and one kinematic token. The shared 69-token state feeds both direct prediction heads and a low-rank projector connected to a frozen language backbone. Aligned, within-task permuted, and no-language conditions change only answer-string alignment or the presence of language loss. Insets show real K-Radar sequence~43 sweeps radar00145--00154 and validation outputs. Proposal centers are hypotheses, not detections; camera and LiDAR are not model inputs.}
\label{fig:framework}
\end{figure*}

\section{Related Work}

\paragraph{Automotive 4D-radar perception.}
Automotive radar work spans benchmark construction, detection, motion, and representation learning. nuScenes established a multimodal driving benchmark with radar, while K-Radar and TJ4DRadSet provide large-scale 4D-radar data across scenes and weather conditions \citep{caesar2020nuscenes,paek2022kradar,zheng2022tj4dradset}. RTNH+ improves radar-tensor detection through two-level CFAR and vertical encoding, while CMFlow learns cross-modally supervised scene flow from temporal radar \citep{kong2023rtnhplus,ding2023hiddengems}. SCKD distills cross-modal knowledge into a radar-only detector, DoppDrive aggregates Doppler evidence over time, and GRT learns transferable raw-radar representations \citep{xu2025sckd,haitman2025doppdrive,huang2025grt}. Sensor-fusion methods combine camera or LiDAR with radar through cross-view fusion, Doppler-aware fusion, or radar-guided geometry enhancement \citep{zhong2025cvfusion,chae2025dopplerfusion,tong2026rel}. RadarSplat uses radar for data synthesis and 3D reconstruction \citep{kung2025radarsplat}. Collectively, this literature improves radar perception or reconstruction rather than language-side attribution. Radar4D-VLM instead uses an RTNH-compatible network only for center proposals, evaluates them by geometric recall rather than detection AP, and exposes temporal radar evidence to language models.

\paragraph{Radar--language interfaces.}
Radar--language systems differ in radar representation and language role. The Radar Spectra--Language Model aligns raw automotive radar spectra with a pretrained vision--language space for retrieval and scene parsing \citep{pushkareva2024radarspectra}. Talk2Radar uses language as an input query for object grounding in a 4D-radar point cloud \citep{guan2025talk2radar}. RLM learns scene-level semantics from simulated radar--caption pairs and spatial supervision \citep{mishra2025rlm}, whereas Weather-Robust Scene Semantics maps K-Radar features through a frozen vision--language model to structured semantic outputs \citep{hamilton2026weather}. RadarLLM maps temporal millimeter-wave point clouds to human-motion representations \citep{lai2026radarllm}. Together, these works establish spectrum--text alignment, language-conditioned grounding, radar captioning, and temporal motion reasoning. Radar4D-VLM instead builds proposal-grounded object tokens, global scene tokens, and an explicit kinematic token from real automotive radar, then asks whether frozen backbones consume this hierarchy, whether outputs depend on radar evidence, and whether aligned supervision improves the shared representation.

\paragraph{Driving and 3D language models.}
General vision--language and 3D-language models provide architectural context. BLIP-2 connects a frozen visual encoder to a frozen language model through a lightweight querying module \citep{li2023blip2}. 3D-LLM and LL3DA extend adapter-based ideas to point-cloud scene understanding and interactive 3D reasoning, while Point Cloud as a Foreign Language treats point-cloud features as language-compatible multimodal inputs \citep{hong2023threedllm,chen2024ll3da,paul2026sage}. PerLA, 3D-GRAND, and situation-aware modeling target local--global perception, grounded instruction data, and observer-centered reasoning \citep{mei2025perla,yang2025threedgrand,yuan2025situation}. Beacon3D shows that object-centric grounding, question answering, and generalization must be evaluated jointly rather than inferred from aggregate answer accuracy \citep{huang2025beacon3d}. These models motivate object-aware tokens, local--global structure, and grounding-specific evaluation for sparse 3D inputs.
Driving-specific language research spans question answering, structured reasoning, and planning. NuScenes-QA defines diagnostic questions over driving scenes, and DriveLM organizes perception, prediction, and planning as graph-structured visual question answering \citep{qian2024nuscenesqa,sima2024drivelm}. DiMA distills multimodal language-model knowledge for driving, OpenDriveVLA connects vision--language representations to end-to-end actions, and Driving with Advice uses a large model as a motion advisor for joint planning \citep{hegde2025dima,zhou2026opendrivevla,wang2026drivingadvice}. DriveBench and VLADBench expose reliability and fine-grained capability gaps in driving VLM evaluation \citep{xie2025drivebench,li2025vladbench}. They motivate compact adapters and disaggregated evaluation under matched tasks; we instead audit a radar-only interface.

\paragraph{Controlled attribution and our position.}
Successful language output establishes interface capability, but not sensor use or representation improvement from the language objective. Multi-task objectives can cooperate or interfere, and shortcut learning can preserve output quality through unintended cues \citep{standley2020tasks,geirhos2020shortcut}. Cross-backbone evaluation tests whether frozen language models can consume the radar tokens. Radar zeroing, shuffling, history reversal, and Doppler removal test sensor and temporal dependence. Finally, aligned, fixed-permutation, and no-language objectives test whether semantic alignment improves matched direct heads while holding radar inputs, task structure, and adaptation budget fixed. Thus, the paper contributes an auditable temporal radar interface with matched objectives and explicit sensor interventions, preventing output compatibility from being mistaken for supervision benefit.

\section{Method}

Radar4D-VLM converts ten consecutive 4D-radar sweeps into a shared physical state that supports both closed-set language inference and language-independent prediction. At time $t$, the current sweep $X_t$ enters a frozen, train-split RTNH-compatible encoder to produce Top-64 proposal centers, proposal features and priors, and a bird's-eye-view (BEV) feature map. All sweeps $X_{t-9:t}$ then supply proposal-aligned temporal and Doppler evidence to trainable object, scene, and kinematic tokenizers. Their outputs are concatenated as
\begin{equation}
\mathbf{H}_t=[\mathbf{o}_{t,1:64};\,\mathbf{k}_t;\,\mathbf{s}_{t,1:4}]
\in\mathbb{R}^{69\times256}.
\end{equation}
Figure~\ref{fig:framework} shows this data flow and the matched supervision conditions.

The same state $\mathbf{H}_t$ drives two synchronized output paths. A learned state query attends over all 69 tokens and feeds direct categorical and velocity heads, providing an auditable path that does not depend on answer tokens. In parallel, a budgeted low-rank projector maps every radar token to the embedding dimension of a frozen language backbone. Sharing the radar state, direct heads, data windows, and optimization budget across conditions lets the experiments distinguish interface compatibility, radar dependence, and benefit from aligned language supervision.

\subsection{Task and Evaluation Endpoints}

Let $X_t=\{p_{t,j}\}_{j=1}^{N_t}$ denote a radar point cloud at time $t$. Given ten sweeps $X_{t-9:t}$, Radar4D-VLM predicts moving-object count and the nearest object's sector, motion, collision risk, category, speed bin, and signed radial velocity. These deterministic label-derived probes form a fixed driving-scene ontology rather than unrestricted language understanding. Radar-only denotes model input and inference: dataset tracks supply supervision but are excluded from the model input.

For task $q$ with label set $\mathcal{Y}_q$, the language path emits a closed-set answer through a frozen backbone, while a learned direct head predicts the same label from the shared radar state. We report a five-task core aggregate over moving count, sector, motion, collision risk, and speed. Category remains auxiliary because the ``other'' class has zero development-validation support. Macro balanced accuracy is computed per task and then averaged, preventing majority classes from dominating the aggregate.

\subsection{Proposal-Grounded Temporal Radar Tokens}

An RTNH-compatible sparse-convolutional model, trained from random initialization on K-Radar sequences 1--40 and selected on sequences 41--48, emits the 64 highest-scoring centers per frame. Fixed-lattice and random controls test whether these centers carry target geometry before downstream training. The proposal outputs are center hypotheses, not detections.

For proposal $i$, the tokenizer combines its frozen semantic feature $\mathbf{f}^{\mathrm{sem}}_i$, normalized center and score $\mathbf{g}_i$, frozen class/box/direction priors $\mathbf{f}^{\mathrm{prior}}_i$, and a learned temporal Doppler descriptor $\mathbf{d}_i$:
\begin{equation}
\mathbf{o}_{t,i}=\operatorname{LN}\!\left(
f_{\mathrm{obj}}\!\left[
\mathbf{f}^{\mathrm{sem}}_i;
f_{\mathrm{geo}}(\mathbf{g}_i);
f_{\mathrm{prior}}(\mathbf{f}^{\mathrm{prior}}_i);
\mathbf{d}_i
\right]\right).
\end{equation}
The Doppler branch associates nearby radar returns with each proposal across $X_{t-9:t}$, exposing displacement, local radial velocity, range slope, and Doppler-alias residuals rather than treating the sweeps as an unordered union.

Global context is retained separately from the proposal stream. A projected BEV grid with positional embeddings forms context $\mathbf{C}_t$, and four learned queries $\mathbf{Q}_s$ produce scene tokens
\begin{equation}
\mathbf{S}_t=\operatorname{LN}\!\left(
\mathbf{Q}_s+\operatorname{MHA}(\mathbf{Q}_s,\mathbf{C}_t,\mathbf{C}_t)
\right).
\end{equation}
For motion summarization, proposal weights combine predicted motion probability, proposal confidence, and range,
\begin{equation}
\alpha_i=\operatorname{softmax}_i\!\left(
\log\sigma(m_i)+\log a_i-\|\mathbf{c}_i\|_2/20
\right),
\end{equation}
where $m_i$, $a_i$, and $\mathbf{c}_i$ are the proposal motion logit, confidence, and center. The kinematic token adds the weighted object state $\sum_i\alpha_i\mathbf{o}_{t,i}$ to an embedding of weighted range, signed and absolute velocity, range slope, local Doppler, alias residual, and moving-probability statistics. A learned state query attends over $\mathbf{H}_t$ to obtain $\mathbf{z}_t$; six direct heads mirror the categorical answer sets, a separate head regresses signed nearest-target radial velocity, and proposal-level heads supervise target match, category, and radial velocity.

\subsection{Frozen-Language Interface and Matched Objectives}

A low-rank radar projector $P$ maps the 69 tokens to a frozen language backbone while keeping trainable projector parameters below 1.2 million. The projected radar prefix $P(\mathbf{H}_t)$ is concatenated with the native token embeddings of a task prompt. For task $q$ and registered answer string $y_q$, the language objective is the mean answer-token negative log-likelihood $\mathcal{L}_{\mathrm{lang}}$. The same interface is evaluated with eight Qwen, Phi, Mistral, Llama, and Gemma backbones; the four Qwen sizes provide a descriptive within-family diagnostic, not a scaling claim.

Training combines the language objective with shared direct and proposal supervision:
\begin{equation}
\begin{aligned}
\mathcal{L}={}&
\lambda_{\mathrm{lang}}\mathcal{L}_{\mathrm{lang}}
+\lambda_{\mathrm{task}}\mathcal{L}_{\mathrm{task}}
+\lambda_{\mathrm{vel}}\mathcal{L}_{\mathrm{vel}}\\
&+\lambda_{\mathrm{mot}}\mathcal{L}_{\mathrm{mot}}
+\lambda_{\mathrm{cat}}\mathcal{L}_{\mathrm{cat}}
+\lambda_{\mathrm{pvel}}\mathcal{L}_{\mathrm{pvel}}.
\end{aligned}
\end{equation}
Here $\mathcal{L}_{\mathrm{task}}$ averages categorical cross-entropy over the active scene tasks, $\mathcal{L}_{\mathrm{vel}}$ is a smooth-$L_1$ loss on signed nearest-target velocity, and the final three terms supervise proposal match, category, and velocity. The aligned condition uses canonical task verbalizers. A fixed within-task derangement preserves task identity, answer-set size, loss weights, and the complete language forward path but maps each label to a wrong answer string. The no-language condition sets $\lambda_{\mathrm{lang}}=0$ and verifies zero projector gradients on every optimization batch. All conditions otherwise retain the same radar architecture and direct heads. Because permutation directly changes answer-token agreement, language value is evaluated on matched direct-head endpoints rather than on the language path itself.

\section{Experiments}

\subsection{Data Isolation and Evaluation Manifest}

We evaluate on K-Radar, a 4D automotive-radar dataset collected across varied weather conditions \citep{paek2022kradar}. The strict split assigns sequences 1--40 to training and sequences 41--48 to development validation and checkpoint selection. Proposal and adapter training can access only sequences 1--48. The reported manifest contains 4,208 ten-frame windows from the eight validation sequences, including a fixed 1,024-window checkpoint-selection subset.

Earlier development received feedback from K-Radar sequences 49--58; we therefore exclude them from the reported evidence rather than treating them as untouched test data. All reported results are explicitly development-validation results. No official nuScenes \citep{caesar2020nuscenes} or locked-holdout evaluation is claimed.

\subsection{Audit Questions and Controlled Comparisons}

Table~\ref{tab:audit-map} states the role and claim boundary of each experiment. The primary language-value audit compares aligned, fixed-permutation, and no-language objectives for Qwen2.5-3B with paired seeds 41--43 on identical windows. Radar-input interventions are separate: the main table reports aligned checkpoints, while the supplement adds direct-head controls for the permuted and no-language objectives.

The cross-family matrix uses the same three seeds and a matched projector budget. Component studies remove temporal history, Doppler, anchor priors, or groups of radar features. Token studies replace the 69 structured tokens by one pooled token or remove the kinematic token. A separate six-cell comparison matches 64 proposal-selected temporal tokens against 64 fixed-grid temporal tokens while holding the four scene tokens, trainable parameter count, prompts, and windows fixed. A deterministic RTNH+physics system predicts the same ontology without an LLM or trained adapter; it is a same-output reference, not a parameter-matched language-path baseline.

\begin{table*}[t]
\centering
\small
\setlength{\tabcolsep}{5pt}
\renewcommand{\arraystretch}{1.16}
\renewcommand{\tabularxcolumn}[1]{m{#1}}
\begin{tabularx}{\textwidth}{
  >{\bfseries\centering\arraybackslash}m{0.14\textwidth}
  >{\raggedright\arraybackslash}m{0.25\textwidth}
  >{\centering\arraybackslash}m{0.19\textwidth}
  >{\raggedright\arraybackslash}X}
\toprule[1.1pt]
\textbf{Question} & \textbf{Controlled comparison} & \textbf{Primary endpoint} & \textbf{Supported interpretation} \\
\midrule[0.8pt]
Proposal geometry
& Learned Top-64 centers vs.\ fixed lattice and uniform random
& Center recall at 4~m; nearest-target distance
& Centers carry target geometry; not detection or AP. \\
\specialrule{0.35pt}{2pt}{2pt}
Interface compatibility
& Same 69-token state across eight frozen backbones
& Language-path core balanced accuracy
& Tokens are consumable across audited families; not a ranking or scaling result. \\
\specialrule{0.35pt}{2pt}{2pt}
Sensor dependence
& Real radar vs.\ zero, shuffle, reverse-history, and zero-Doppler inputs
& Language- and direct-path score change
& Aligned checkpoints use radar content and temporal order; not evidence of language value. \\
\specialrule{0.35pt}{2pt}{2pt}
Language value
& Aligned vs.\ fixed permutation vs.\ no language on matched seeds and windows
& Equal-sequence paired direct-head effect
& Tests whether aligned language supervision improves shared radar utility. \\
\specialrule{0.35pt}{2pt}{2pt}
Structure and deployment
& Ten vs.\ one sweep; 69 tokens vs.\ pooling or token removal; pruned no-LLM export
& Path deltas and export drift envelope
& Bounds component utility and direct deployment; not causal proposal isolation. \\
\bottomrule[1.1pt]
\end{tabularx}
\caption{Evaluation map. Each comparison answers one question and carries an explicit claim boundary; results from one row are not used as substitutes for another.}
\label{tab:audit-map}
\vspace{-6pt}
\end{table*}

\subsection{Metrics and Statistical Analysis}

The formal nearest-category order is none, sedan, large vehicle, pedestrian, and other, with development-validation supports 1,371/1,993/524/320/0. The largest imbalance is collision risk: 3,355 of 4,208 windows (79.7\%) are low risk. Full per-class counts are reported in the supplement. Balanced accuracy averages classwise recall within each task before task-level aggregation; signed-velocity error is reported separately, preventing regression quality from being folded into the categorical core score under the same manifest.

Condition means use sample-pooled atomic-confusion metrics, so sequences contribute in proportion to their window support. Primary language-control contrasts instead pair conditions within seed, compute sequence-level differences, and weight each of the eight sequences equally. Crossed seed--sequence bootstrap intervals resample both axes. Windows are never treated as independent replicates. With three seeds, the comparisons remain descriptive because an exact two-sided seed sign-flip test cannot attain conventional significance.

\subsection{Matched Language-Supervision Audit}

The primary result is a bounded non-observation of language-supervision benefit. Table~\ref{tab:language-control} reports the pre-specified equal-sequence paired effects, which pair objectives within each seed and give each validation sequence equal weight. Aligned minus permuted supervision changes direct-head core balanced accuracy by $-0.0052$ (95\% crossed seed--sequence interval [$-0.0305$, 0.0220]); aligned minus no language changes it by $-0.0024$ ([$-0.0231$, 0.0188]). Neither contrast supports a stable gain in direct-head predictive utility. Sample-pooled condition means are reported in the supplement.

This result is descriptive rather than an equivalence claim: the audit contains one fixed verbalizer permutation and three training seeds, for which an exact two-sided seed sign-flip test cannot attain conventional significance. The interval upper endpoints are 0.0220 and 0.0188 balanced-accuracy units, so positive effects up to approximately 2.20 and 1.88 percentage points remain compatible with the data. This is a bounded non-observation, not a practical-equivalence claim.

\begin{table}[t]
\centering
\small
\setlength{\tabcolsep}{4pt}
\begin{tabular}{lcc}
\toprule
Primary contrast & Paired effect & 95\% interval \\
\midrule
Aligned $-$ permuted & $-0.0052$ & [$-0.0305$, 0.0220] \\
Aligned $-$ no language & $-0.0024$ & [$-0.0231$, 0.0188] \\
\bottomrule
\end{tabular}
\caption{Primary equal-sequence paired effects on Qwen2.5-3B direct-head core balanced accuracy (seeds 41--43). Intervals use crossed seed--sequence bootstrap resampling.}
\label{tab:language-control}
\vspace{-6pt}
\end{table}

\begin{table}[t]
\centering
\scriptsize
\setlength{\tabcolsep}{2.6pt}
\begin{tabularx}{\linewidth}{
  >{\raggedright\arraybackslash}X
  >{\centering\arraybackslash}p{0.34\linewidth}
  >{\centering\arraybackslash}p{0.34\linewidth}}
\toprule
Real minus control & Language path $\Delta$ [95\% interval] & Direct path $\Delta$ [95\% interval] \\
\midrule
Strict-zero radar
& 0.1215 [0.0618, 0.1947]
& 0.1370 [0.0724, 0.2107] \\
Scene-local shuffle
& 0.1532 [0.0704, 0.2639]
& 0.1703 [0.0842, 0.2725] \\
Reversed history
& 0.0249 [0.0073, 0.0433]
& 0.0277 [0.0097, 0.0464] \\
Zero Doppler
& 0.0101 [$-0.0077$, 0.0296]
& 0.0106 [$-0.0051$, 0.0280] \\
\bottomrule
\end{tabularx}
\caption{Aligned Qwen2.5-3B real-minus-control effects (seeds 41--43). Positive values indicate lower accuracy under the named input control; intervals use crossed seed--sequence bootstrap resampling.}
\label{tab:radar-controls-main}
\vspace{-6pt}
\end{table}

\subsection{Radar Dependence and Temporal Order}

The language-value null is not explained by a failed task or an aligned model that ignores radar. Learned direct-head core scores near 0.50 exceed the ontology's macro chance reference of approximately 0.283 and the deterministic RTNH+physics reference of 0.3401. Input controls use only the aligned Qwen2.5-3B checkpoints for seeds 41--43. On the language path, strict-zero radar reduces core balanced accuracy by 0.1215 (crossed seed--sequence 95\% interval [0.0618, 0.1947]), scene-local shuffling reduces it by 0.1532 [0.0704, 0.2639], and reversing the ten-sweep history reduces it by 0.0249 [0.0073, 0.0433]. The corresponding direct-path reductions are 0.1370, 0.1703, and 0.0277. The zero-Doppler language-path effect is 0.0101 with an interval crossing zero.

These controls show that the aligned pathway uses radar content and temporal order. Supplementary direct-head controls show the same strict-zero, scene-local-shuffle, repeated-current, and reversed-history directions under the permuted and no-language objectives. This objective-robust sensor dependence does not convert the bounded language-value null into evidence of benefit caused by aligned language supervision.

This distinction is the central interpretation of the experiment: successful language output and sensitivity to radar interventions establish interface capability and sensor use, but neither is evidence that aligned language supervision improves the shared radar representation.

\subsection{Qualitative and Attribution Evidence}

\begin{figure*}[t]
\centering
\includegraphics[width=\textwidth]{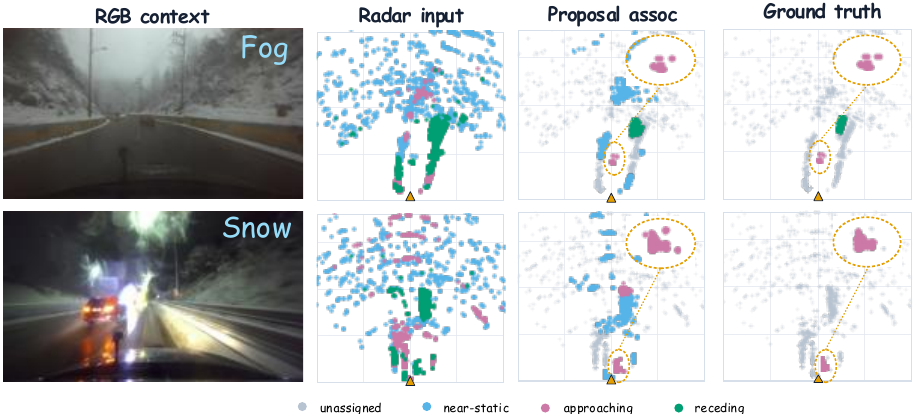}
\caption{Proposal-to-return association in fog and snow. RGB is synchronized context only and is not a model input.}
\label{fig:audit-evidence-chain}
\end{figure*}

\begin{figure}[t]
\centering
\includegraphics[width=\columnwidth]{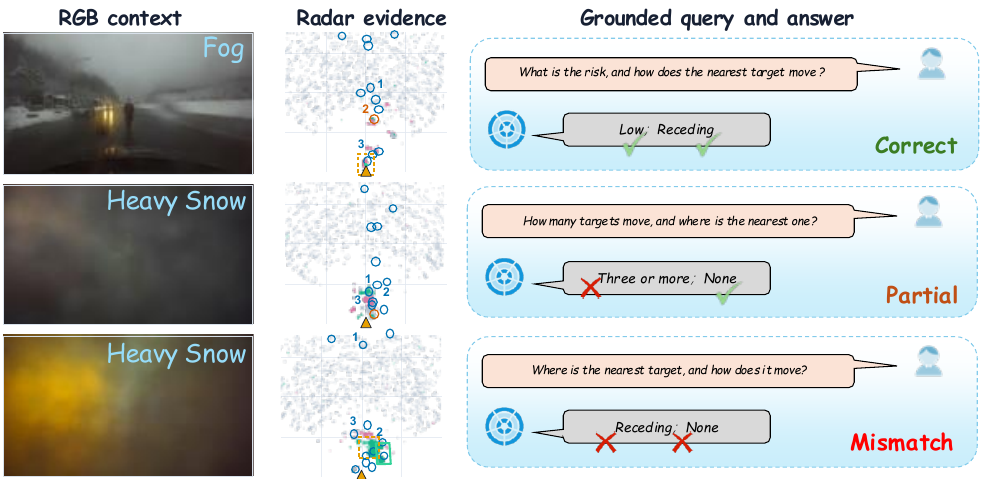}
\caption{Grounded Q\&A cases spanning correct, partial, and mismatched outputs.}
\label{fig:main-qualitative}
\end{figure}

Figures~\ref{fig:audit-evidence-chain} and~\ref{fig:main-qualitative} connect proposal-level geometry to the final ontology outputs. In Figure~\ref{fig:audit-evidence-chain}, the fog and snow examples retain the same progression from raw radar returns to proposal association and track-derived ground truth. ``Proposal assoc.'' assigns current-frame returns to frozen bundle proposals within 3.6~m, with colors separating unassigned, near-static, approaching, and receding returns. In both rows, the highlighted proposal collects a spatially coherent group of approaching returns around the tracked vehicle while leaving most background and near-static returns unassigned. Agreement in position and motion sign indicates that the frozen proposal interface supplies usable target evidence under adverse weather; it is not predicted scene flow, point-level semantic segmentation, object detection, or AP.

Figure~\ref{fig:main-qualitative} exposes how this evidence appears in a multi-field language response. The fog case answers both collision risk and nearest-target motion correctly. The first heavy-snow case preserves the nearest-location answer but overcounts moving targets, whereas the final case misses both requested fields. Thus, a fluent response can still be only partially correct, and the ontology fields should be scored separately. RGB is synchronized context only in both figures. The three language outputs are fixed Qwen2.5-0.5B seed-42 examples shown only for qualitative inspection, not an aggregate performance estimate.

\begin{figure}[t]
\centering
\includegraphics[width=\columnwidth]{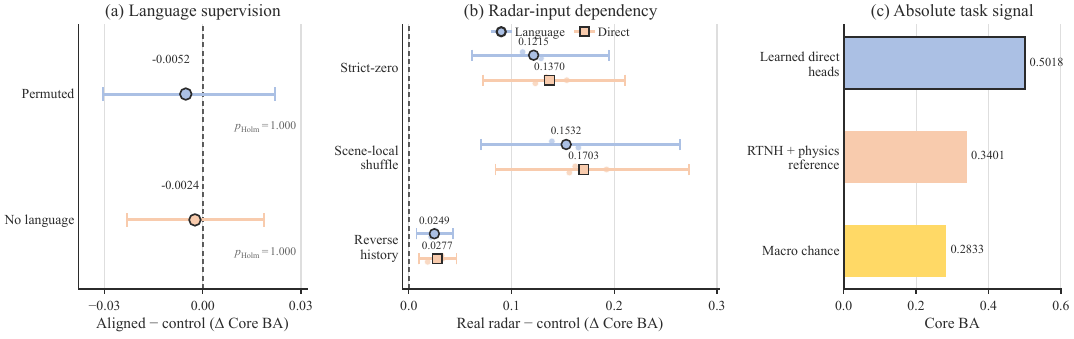}
\caption{Controlled attribution on K-Radar development validation. Error bars show crossed seed--sequence 95\% intervals.}
\label{fig:controlled-attribution-summary}
\end{figure}

Figure~\ref{fig:controlled-attribution-summary} makes this separation explicit. Panel (a) places both aligned-minus-control effects slightly below zero, with both crossed seed--sequence intervals spanning zero and Holm-adjusted $p=1.000$; with three seeds, these adjusted values remain descriptive. Accordingly, the panel supports a bounded non-observation rather than evidence that aligned supervision is harmful. Panel (b), in contrast, shows positive real-minus-control drops for strict-zero radar, scene-local shuffling, and reversed history on both output paths. Panel (c) provides a third check: the learned direct heads reach 0.5018 core balanced accuracy, above the RTNH+physics reference (0.3401) and macro chance (0.2833).

The three panels answer complementary questions rather than forming a single ranking or composite score. Panel (a) tests benefit from aligned language supervision, panel (b) tests whether the aligned checkpoints use radar content, and panel (c) verifies that the task signal is nontrivial. This decomposition keeps predictive competence separate from the causal interpretation of the training objective. Their joint reading explains why a language-value null can coexist with strong radar dependence and above-reference direct prediction.

\subsection{Backbone Compatibility and Structural Evidence}

The attribution result does not imply that frozen language models cannot consume the radar tokens. Across the complete 24-cell language-path matrix, core balanced accuracy lies in a narrow 0.4860--0.4995 range, all sequence-cluster intervals overlap, the four Qwen sizes show no monotone trend, and non-Qwen entries occupy the same range. The supplement reports the eight aggregate rows, intervals, and all raw seed metrics. The evidence therefore supports compatibility of the 69-token interface across the audited backbones, but not scaling, a best family, or language-model necessity.

The deterministic reference's all-six-task balanced accuracy is 0.3359 with a sequence-cluster 95\% interval of [0.3040, 0.3694]. Because it uses fixed rules rather than the learned radar tokenizer, it is not a matched comparison of LLM necessity.

\paragraph{Proposal geometry and representation structure.}

The proposal source is geometrically non-degenerate. On 4,289 validation frames with 4,062 vehicle-region targets, the selected checkpoint reaches Top-64 center recall of 0.9813 at 4~m, compared with 0.9173 for a fixed lattice and 0.7530 for uniform random proposals (20-seed SD 0.0071). Its mean nearest-target distance is 1.057~m versus 2.426~m for the lattice. These are center-geometry diagnostics, not detections or detector AP.

\begin{figure}[t]
\centering
\includegraphics[width=\columnwidth]{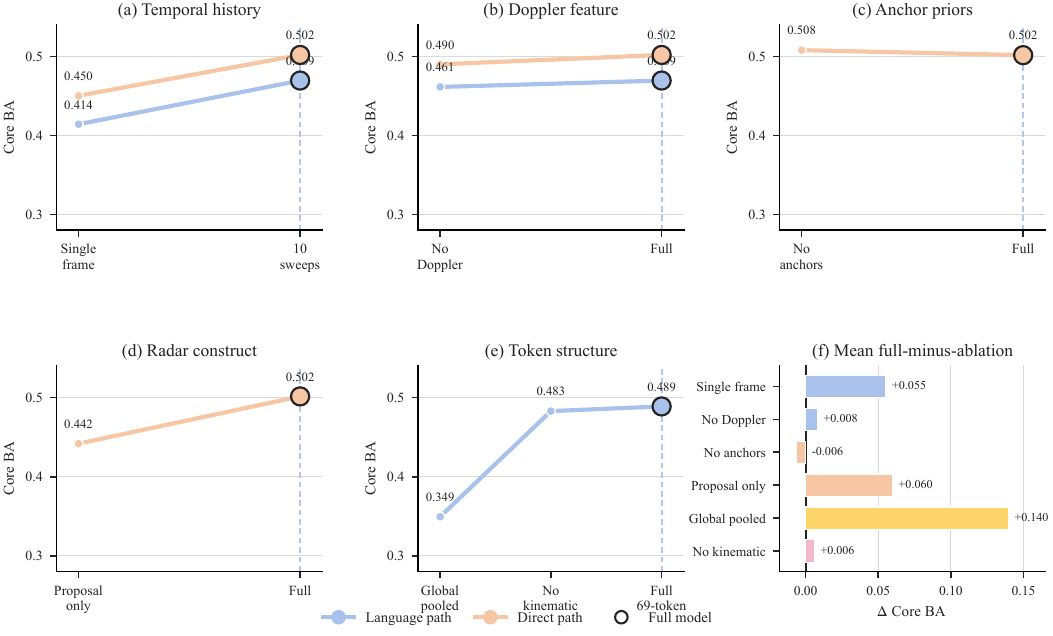}
\caption{Structural ablations on K-Radar development validation. Panel (f) reports mean full-minus-ablation changes in core balanced accuracy.}
\label{fig:structural-ablation}
\end{figure}

Figure~\ref{fig:structural-ablation} identifies which interface choices carry the largest task signal using seeds 41--43 and the same 4,208 development-validation windows per cell. Panels (a--b) separate temporal accumulation from Doppler. Replacing ten sweeps by the current sweep lowers core balanced accuracy by 0.0552 on the language path and 0.0516 on the direct path, with the same direction for all seeds. Removing Doppler produces a smaller direct-path reduction of 0.0119; the language-path change is only 0.0081 in full-minus-ablation magnitude and has mixed seed directions. Temporal history is therefore the more consistent of these two cues.

Panels (c--f) test representation structure. Removing anchor priors slightly improves the no-language direct score (0.5078 versus 0.5015), while retaining only proposal features and center/score geometry lowers it to 0.4418. On the language path, global pooling produces the largest degradation, from 0.4889 to 0.3493, whereas removing only the kinematic token yields 0.4829 with mixed seed-level differences. The proposal-only and pooling comparisons remain compound interventions. The supplement therefore adds an equal-count control: 64 proposal-selected temporal tokens exceed 64 fixed-grid temporal tokens by 0.0456 core balanced-accuracy units under equal-sequence pairing, with all three seed directions positive and a 95\% sequence-cluster interval of [0.0241, 0.0660].

\subsection{LLM-Free Direct Inference}

The no-language estimators enable deployment without a language model, tokenizer, or projector, removing the language stack from direct inference. Exact tensor pruning preserves retained direct-model tensors, and the three no-LLM exports achieve mean core balanced accuracy of 0.5014 (seed SD 0.0229). Fresh evaluation does not reproduce parent raw traces exactly: each export satisfies only the registered same-checkpoint GPU-nondeterminism envelope. This distinction separates architectural removal from numerical identity. Direct prediction remains operational after excising the LLM components, with output differences bounded by the audited envelope. These exports demonstrate a simplified LLM-free deployment path, not bitwise prediction preservation, additional training replicates, or measured improvements in runtime or memory efficiency.

\subsection{Limitations}

The equal-count proposal-versus-grid control isolates token location and aggregation more closely than pooling, but remains a single Qwen2.5-3B development-validation design with the same proposal auxiliary branch. Results cover six sensor-observable questions; risk is low-class dominated, ``other'' lacks support, and prior feedback on sequences 49--58 precludes an untouched test estimate. External generalization and state-of-the-art performance are untested. The primary language audit still uses one fixed verbalizer permutation and three seeds; three additional derangements reveal mapping sensitivity but remain ineligible for confirmatory seed inference. Temporal and Doppler findings remain conditional on the fixed proposal interface. We therefore claim only no observed language-supervision benefit in this setting, alongside evidence of temporal-radar use.

\section{Conclusion}

This controlled attribution audit separates interface compatibility, sensor dependence, and benefit from aligned language supervision. Within a closed K-Radar development-validation ontology, the first two claims are supported, whereas aligned supervision yields no stable gain in co-adapted direct-head predictive utility over permuted or no-language objectives. Exact tensor pruning further enables an LLM-free direct export with envelope-bounded drift. More broadly, these three claims should be tested and reported separately.

\bibliography{references}

\clearpage
\appendix

\section{Scope of This Supplement}

The main paper presents the claims and headline comparisons. This supplement provides the material
needed to audit those claims without repeating the main narrative: complete label supports, formal
optimization settings, proposal-geometry diagnostics, aggregate and seed-level cross-backbone results,
raw supporting ablations, and the full results of three validation-only experiment groups completed after
the main analysis was frozen. Those three groups are (i) six matched proposal/grid training cells,
(ii) six checkpoint-evaluation cells covering two objectives and five input interventions, and (iii) nine
cells spanning three additional fixed verbalizer permutations. BA denotes macro balanced accuracy over
the five-task core endpoint unless stated otherwise.

\section{Common Protocol and Statistical Boundary}

All three experiment groups use K-Radar sequences 1--40 for training and sequences 41--48 for
development validation. Every reported cell is evaluated on the same ordered 4,208 ten-sweep windows.
Training seeds are 41, 42, and 43. The newly executed jobs were restricted to the validation split and
did not create or read a new test.json file. K-Radar sequences 49--58 were not accessed by these
experiments. Results therefore remain development-validation evidence rather than independent-test or
external-transfer evidence.

For the matched tokenization experiment, the primary contrast computes the proposal-minus-grid
difference within each seed and validation sequence, then gives the eight sequences equal weight.
Sequence-cluster intervals use 5,000 paired bootstrap samples conditional on the fixed seeds. Crossed
intervals resample both sequences and training seeds. Information-control intervals use the same
5,000-sample crossed procedure. Additional-verbalizer intervals are conditional on the three fixed seeds.
Window-level predictions are never treated as independent replicates.

\section{Formal Settings and Label Supports}

Table~\ref{tab:formal-settings} records the shared launcher settings that are not expanded in the main
paper. Model-specific projector ranks and tokenizer hashes remain fixed by the corresponding run
manifests. The first epoch warms up the radar and direct components without language loss. Epochs two
through eight activate the registered language objective, except in the no-language condition.

\begin{table}[t]
\centering
\scriptsize
\setlength{\tabcolsep}{3pt}
\begin{tabular}{ll}
\toprule
Setting & Frozen value \\
\midrule
History / tokens & 10 sweeps / 64 object + 4 scene + 1 kinematic \\
Radar points / neighborhood & 10,240 / 128 \\
Hidden dimension & 256 \\
Training seeds & 41, 42, 43 \\
Epochs / samples per epoch & 8 / 8,192 \\
Train / evaluation batch size & 2 / 2 \\
Optimizer & AdamW, $3\times10^{-4}$, weight decay 0.01 \\
Projector budget & at most 1.2M trainable parameters \\
Checkpoint-selection subset & 1,024 validation windows \\
Reported denominator & 4,208 validation windows \\
Prompt serialization & native tokenizer chat template \\
\bottomrule
\end{tabular}
\caption{Shared formal adapter settings. The immutable per-run manifests remain authoritative.}
\label{tab:formal-settings}
\end{table}

Table~\ref{tab:label-support} reports the complete development-validation label denominator. The
``other'' category has zero support and nearest category is therefore excluded from the five-task core
aggregate. Balanced accuracy averages recall over classes with nonzero support and does not inherit the
largest-class prevalence.

\begin{table*}[t]
\centering
\scriptsize
\setlength{\tabcolsep}{4pt}
\begin{tabular}{p{0.15\textwidth}p{0.54\textwidth}rr}
\toprule
Task & Class supports in ontology order & Largest class & Chance BA \\
\midrule
Moving count & none 1,371; one 1,696; two 904; $\geq3$ 237 & 40.3\% & 0.250 \\
Nearest sector & none 1,371; left 828; ahead 1,649; right 360 & 39.2\% & 0.250 \\
Nearest motion & none 1,371; approaching 1,699; receding 1,138 & 40.4\% & 0.333 \\
Collision risk & low 3,355; medium 435; high 418 & 79.7\% & 0.333 \\
Nearest category$^\dagger$ & none 1,371; sedan 1,993; large vehicle 524; pedestrian 320; other 0 & 47.4\% & 0.250 \\
Nearest speed & none 1,371; slow 2,048; medium 378; fast 411 & 48.7\% & 0.250 \\
\bottomrule
\end{tabular}
\caption{Source-bound class supports on the common 4,208-window validation manifest. ``Largest class''
is descriptive prevalence. Chance BA is $1/C$ over the $C$ classes with nonzero support.
$^\dagger$Auxiliary task, excluded from the five-task core aggregate.}
\label{tab:label-support}
\end{table*}

\section{Proposal Geometry Beyond the Headline Result}

The strict proposal checkpoint was selected by Top-64 vehicle-region recall at 4~m on sequences
41--48, with validation loss used only for an exact tie. The diagnostic contains 4,289 validation
frames and 4,062 vehicle-region targets. Table~\ref{tab:proposal-grid} gives the complete
proposal-count/radius grid. These are center-hypothesis diagnostics rather than detections or AP.

\begin{table*}[t]
\centering
\scriptsize
\setlength{\tabcolsep}{5pt}
\begin{tabular}{rrrrr@{\qquad}rrrrr}
\toprule
$K$ & Radius & Selected & Lattice & Random & $K$ & Radius & Selected & Lattice & Random \\
\midrule
8  & 0.5 & 0.4030 & 0.0034 & $0.0024\pm0.0008$ & 16 & 0.5 & 0.4116 & 0.0057 & $0.0055\pm0.0011$ \\
8  & 1.0 & 0.6435 & 0.0113 & $0.0103\pm0.0019$ & 16 & 1.0 & 0.6620 & 0.0158 & $0.0211\pm0.0019$ \\
8  & 2.0 & 0.7720 & 0.0411 & $0.0414\pm0.0023$ & 16 & 2.0 & 0.8067 & 0.0618 & $0.0835\pm0.0044$ \\
8  & 4.0 & 0.9101 & 0.1691 & $0.1582\pm0.0075$ & 16 & 4.0 & 0.9540 & 0.2797 & $0.2963\pm0.0071$ \\
32 & 0.5 & 0.4136 & 0.0133 & $0.0107\pm0.0013$ & 64 & 0.5 & 0.4136 & 0.0234 & $0.0219\pm0.0017$ \\
32 & 1.0 & 0.6669 & 0.0446 & $0.0424\pm0.0025$ & 64 & 1.0 & 0.6679 & 0.0963 & $0.0844\pm0.0051$ \\
32 & 2.0 & 0.8176 & 0.1800 & $0.1608\pm0.0031$ & 64 & 2.0 & 0.8230 & 0.3774 & $0.2952\pm0.0082$ \\
32 & 4.0 & 0.9734 & 0.6839 & $0.5042\pm0.0069$ & 64 & 4.0 & 0.9813 & 0.9173 & $0.7530\pm0.0071$ \\
\bottomrule
\end{tabular}
\caption{Complete proposal recall grid. Radius is in meters, and random values are mean $\pm$ SD across
20 fixed random seeds.}
\label{tab:proposal-grid}
\end{table*}

\section{Raw Results for the Main Attribution Audits}

The main paper uses equal-sequence paired contrasts for language attribution. Table~\ref{tab:pooled-language}
reports the secondary sample-pooled condition means reconstructed from atomic confusion matrices. These
means weight sequences by their number of windows and therefore do not replace the primary estimand.

\begin{table}[t]
\centering
\scriptsize
\setlength{\tabcolsep}{4pt}
\begin{tabular}{lrrr}
\toprule
Training condition & Direct BA & Direct F1 & Seeds \\
\midrule
Aligned language & 0.5018 & 0.4855 & 3 \\
Fixed verbalizer permutation & 0.5008 & 0.4832 & 3 \\
No language objective & 0.5015 & 0.4786 & 3 \\
\bottomrule
\end{tabular}
\caption{Sample-pooled direct-head condition means on 4,208 windows per seed.}
\label{tab:pooled-language}
\end{table}

Tables~\ref{tab:cross-family-aggregate} and~\ref{tab:cross-family-seeds} provide the full results behind
the cross-family compatibility claim. All 24 cells use the same ordered validation manifest and the same
projector-parameter ceiling. The analysis supports compatibility across the audited backbones, not a
family ranking or scaling law.

\begin{table}[t]
\centering
\scriptsize
\setlength{\tabcolsep}{3pt}
\begin{tabular}{lrrr}
\toprule
Frozen backbone & BA & 95\% seq. CI & F1 \\
\midrule
Qwen2.5-0.5B & 0.4868 & [0.4157, 0.5500] & 0.4785 \\
Qwen2.5-1.5B & 0.4940 & [0.4201, 0.5556] & 0.4862 \\
Qwen2.5-3B & 0.4889 & [0.4132, 0.5479] & 0.4698 \\
Qwen2.5-7B & 0.4860 & [0.4073, 0.5530] & 0.4665 \\
Phi-3.5-mini & 0.4866 & [0.4189, 0.5514] & 0.4724 \\
Mistral-7B & 0.4951 & [0.4199, 0.5554] & 0.4814 \\
Llama-3.2-3B & 0.4980 & [0.4265, 0.5553] & 0.4874 \\
Gemma-2-2B-IT & 0.4995 & [0.4322, 0.5570] & 0.4898 \\
\bottomrule
\end{tabular}
\caption{Language-path aggregate results. Intervals resample validation sequences conditional on the
three fixed seeds.}
\label{tab:cross-family-aggregate}
\end{table}

\begin{table}[H]
\centering
\scriptsize
\setlength{\tabcolsep}{3.2pt}
\begin{tabular}{lrrr}
\toprule
\multicolumn{4}{c}{Core BA} \\
\cmidrule(lr){1-4}
Frozen backbone & S41 & S42 & S43 \\
\midrule
Qwen2.5-0.5B  & 0.4964 & 0.4971 & 0.4670 \\
Qwen2.5-1.5B  & 0.4946 & 0.4922 & 0.4953 \\
Qwen2.5-3B    & 0.4820 & 0.4911 & 0.4935 \\
Qwen2.5-7B    & 0.4733 & 0.5036 & 0.4812 \\
Phi-3.5-mini  & 0.4762 & 0.4954 & 0.4881 \\
Mistral-7B    & 0.4751 & 0.5309 & 0.4794 \\
Llama-3.2-3B  & 0.4898 & 0.5077 & 0.4966 \\
Gemma-2-2B-IT & 0.4738 & 0.5302 & 0.4946 \\
\addlinespace
\multicolumn{4}{c}{Core macro-F1} \\
\cmidrule(lr){1-4}
Frozen backbone & S41 & S42 & S43 \\
\midrule
Qwen2.5-0.5B  & 0.4822 & 0.4893 & 0.4511 \\
Qwen2.5-1.5B  & 0.4777 & 0.4854 & 0.4902 \\
Qwen2.5-3B    & 0.4593 & 0.4631 & 0.4746 \\
Qwen2.5-7B    & 0.4392 & 0.4852 & 0.4712 \\
Phi-3.5-mini  & 0.4525 & 0.4677 & 0.4823 \\
Mistral-7B    & 0.4414 & 0.5239 & 0.4586 \\
Llama-3.2-3B  & 0.4715 & 0.4907 & 0.4924 \\
Gemma-2-2B-IT & 0.4593 & 0.5196 & 0.4772 \\
\bottomrule
\end{tabular}
\caption{All seed-level language-path results. Every cell contains 4,208 validation windows.}
\label{tab:cross-family-seeds}
\end{table}

Across the eight backbones, the aggregate BA occupies a narrow 0.4860--0.4995 range
(Table~\ref{tab:cross-family-aggregate}), while Table~\ref{tab:cross-family-seeds} shows that
seed variation can be larger than the difference between backbone means. The two tables therefore
support the intended compatibility statement: the same bounded radar-to-language interface trains and
evaluates across all audited families. They do not provide evidence for ranking model families or for a
monotonic benefit from increasing language-model size.

\section{Raw Supporting Ablations and References}

The following tables disclose the seed-level values summarized in the main paper. They are retained here
because the intervention scope matters: global pooling changes token count and structure jointly, while
the proposal-features-plus-geometry condition removes multiple information sources.

\begin{table}[H]
\centering
\scriptsize
\setlength{\tabcolsep}{3.5pt}
\begin{tabular}{lrrrr}
\toprule
\multicolumn{5}{c}{Language-path BA} \\
\cmidrule(lr){1-5}
Condition & S41 & S42 & S43 & Mean \\
\midrule
Full temporal model & 0.4894 & 0.4709 & 0.4478 & 0.4694 \\
Single frame & 0.3774 & 0.4335 & 0.4315 & 0.4141 \\
Zero Doppler & 0.4438 & 0.4730 & 0.4670 & 0.4613 \\
\addlinespace
\multicolumn{5}{c}{Direct-path BA} \\
\cmidrule(lr){1-5}
Condition & S41 & S42 & S43 & Mean \\
\midrule
Full temporal model & 0.4901 & 0.5260 & 0.4892 & 0.5018 \\
Single frame & 0.4413 & 0.4517 & 0.4576 & 0.4502 \\
Zero Doppler & 0.4795 & 0.5216 & 0.4686 & 0.4899 \\
\bottomrule
\end{tabular}
\caption{Component-suite core BA on the common validation manifest.}
\label{tab:component-seeds}
\end{table}

Relative to the full temporal model, the single-frame condition reduces mean BA by 0.0553 on the
language path and by 0.0516 on the direct path (Table~\ref{tab:component-seeds}). Zeroing Doppler has
a smaller effect, reducing the corresponding means by 0.0081 and 0.0119. Thus, the clearest component
signal is the value of multi-sweep temporal context; the Doppler result is directionally weaker and is not
used as an independent causal claim.

\begin{table}[H]
\centering
\scriptsize
\setlength{\tabcolsep}{3pt}
\begin{tabular}{lrrrr}
\toprule
No-language construct & S41 & S42 & S43 & Mean \\
\midrule
Full information & 0.4821 & 0.5266 & 0.4959 & 0.5015 \\
No anchor priors & 0.4872 & 0.5410 & 0.4954 & 0.5078 \\
Proposal features + geometry only & 0.4373 & 0.4477 & 0.4403 & 0.4418 \\
\bottomrule
\end{tabular}
\caption{Direct-path construct audit. The last row is a compound removal.}
\label{tab:construct-seeds}
\end{table}

Removing anchor priors alone does not reduce the mean direct-path BA in Table~\ref{tab:construct-seeds}
(0.5078 versus 0.5015). In contrast, retaining only proposal features and geometry yields 0.4418. Because
that last condition removes several information sources at once, it establishes that proposal features and
geometry are insufficient by themselves; it does not isolate the contribution of any one removed input.

\begin{table}[H]
\centering
\scriptsize
\setlength{\tabcolsep}{3pt}
\begin{tabular}{lrrrr}
\toprule
Language-path token design & S41 & S42 & S43 & Mean \\
\midrule
Full 69-token interface & 0.4820 & 0.4911 & 0.4935 & 0.4889 \\
One pooled BEV token & 0.3675 & 0.3942 & 0.2864 & 0.3493 \\
No kinematic token & 0.4913 & 0.4755 & 0.4819 & 0.4829 \\
\bottomrule
\end{tabular}
\caption{Token-design audit. Pooling changes token count and token structure jointly.}
\label{tab:token-seeds}
\end{table}

Replacing the 69-token interface with one pooled BEV token lowers mean language-path BA from 0.4889
to 0.3493, whereas removing only the kinematic token changes the mean to 0.4829
(Table~\ref{tab:token-seeds}). This comparison supports preserving distributed object-level tokens, but it
does not attribute the pooled-token loss solely to token count because pooling also changes token structure.

Exact tensor pruning of the no-language checkpoints produces direct-only exports that construct no LLM,
tokenizer, or language projector. Fresh evaluation preserves ordered-window identity and passes the
registered same-checkpoint GPU nondeterminism envelope, but the raw traces are not exactly identical to
the parent traces. Table~\ref{tab:direct-export} therefore describes a deployment transformation rather
than new training replicates.

\begin{table}[H]
\centering
\scriptsize
\begin{tabular}{rrrr}
\toprule
Seed & Core BA & Runtime (s) & Peak GPU (GB) \\
\midrule
41 & 0.4817 & 59.11 & 0.561 \\
42 & 0.5265 & 59.13 & 0.561 \\
43 & 0.4959 & 59.35 & 0.561 \\
\bottomrule
\end{tabular}
\caption{Envelope-verified direct-only exports on 4,208 validation windows. Runtime and memory are
hardware-specific diagnostics.}
\label{tab:direct-export}
\end{table}

Table~\ref{tab:direct-export} is a deployment check rather than another accuracy experiment. The pruned
models retain the parent checkpoints' direct-path BA within the registered nondeterminism envelope while
evaluating all 4,208 windows in about 59 seconds with 0.561~GB peak GPU memory. These hardware-specific
numbers show that the direct predictor can be exported without constructing the frozen LLM; they are not
a cross-platform speed benchmark.

The deterministic RTNH+physics reference uses the train-split proposal checkpoint, analytic range slope,
frozen ontology thresholds, and a train-only category prior. It contains no LLM or trained adapter and is
not parameter matched to Radar4D-VLM.

\begin{table}[H]
\centering
\scriptsize
\setlength{\tabcolsep}{3pt}
\begin{tabular}{lrc}
\toprule
Output & BA & Sequence 95\% CI \\
\midrule
Moving count & 0.2889 & [0.2641, 0.3173] \\
Nearest sector & 0.3539 & [0.2988, 0.4237] \\
Nearest motion & 0.4117 & [0.3636, 0.4559] \\
Collision risk & 0.3440 & [0.3339, 0.3533] \\
Nearest category$^\dagger$ & 0.3148 & [0.2866, 0.4098] \\
Nearest speed & 0.3020 & [0.2670, 0.3321] \\
All six tasks & 0.3359 & [0.3040, 0.3694] \\
\bottomrule
\end{tabular}
\caption{Deterministic same-output reference. The five-task core mean is 0.3401.
$^\dagger$Auxiliary task.}
\label{tab:physics-reference}
\end{table}

The deterministic reference reaches 0.3401 five-task core BA (Table~\ref{tab:physics-reference}), below
the learned direct-path results reported above. Its purpose is to anchor the task against a transparent
same-output procedure, not to serve as a parameter-matched competitor. The per-output values also show
that the analytic rules do not collapse to the largest-class prevalence reported in
Table~\ref{tab:label-support}.

\section{Equal-Token Proposal-versus-Grid Control}

The comparison uses the same frozen Qwen2.5-3B backbone and native-chat prompts in both conditions.
Radar4D-VLM supplies 64 proposal-selected temporal object tokens. The control samples 64 temporal tokens
at the centers of a fixed $8\times8$ region-of-interest grid. Both append the same four scene tokens and
disable the kinematic token, yielding 68 radar-prefix tokens. They retain the same proposal auxiliary
branch, shared modules, initialization within seed, windows, optimizer, checkpoint rule, and projector
budget. Each condition has 1,197,568 trainable semantic-path parameters and 3,663,391 total
trainable parameters.

\begin{table}[H]
\centering
\scriptsize
\setlength{\tabcolsep}{2.5pt}
\begin{tabular}{llrrrr}
\toprule
Endpoint & Tokens & S41 & S42 & S43 & Mean \\
\midrule
Language path & Proposals & 0.4792 & 0.4693 & 0.4971 & 0.4819 \\
Language path & $8\times8$ grid & 0.4148 & 0.4495 & 0.4434 & 0.4359 \\
\addlinespace
Direct head & Proposals & 0.4819 & 0.4882 & 0.5090 & 0.4930 \\
Direct head & $8\times8$ grid & 0.4371 & 0.4737 & 0.4510 & 0.4539 \\
\bottomrule
\end{tabular}
\caption{Raw seed-level core BA for the matched 68-token experiment. Means pool each complete validation cell.}
\label{tab:matched-raw}
\end{table}

Table~\ref{tab:matched-raw} gives the unaggregated sanity check: proposal tokens outperform the equal-count
grid for every seed on both endpoints. The pooled mean gaps are 0.0460 BA on the language path and 0.0391
on the direct head. These values summarize complete cells but still weight validation sequences by their
window counts, so they are not the registered primary effect.

\begin{table}[H]
\centering
\scriptsize
\setlength{\tabcolsep}{3pt}
\begin{tabular}{lrcc}
\toprule
Endpoint & Mean $\Delta$ & Seq. 95\% CI & Crossed 95\% CI \\
\midrule
Language path & 0.0456 & [0.0241, 0.0660] & [0.0181, 0.0736] \\
Direct head & 0.0397 & [0.0221, 0.0552] & [0.0190, 0.0613] \\
\bottomrule
\end{tabular}
\caption{Equal-sequence paired proposal-minus-grid effects. All three seed-level effects are positive,
and the pre-registered language-path gate passes.}
\label{tab:matched-effects}
\end{table}

Table~\ref{tab:matched-effects} reports the primary equal-sequence estimand. The proposal-minus-grid
effect is 0.0456 BA on the language path and 0.0397 on the direct head; both the sequence-cluster and
crossed intervals exclude zero. Agreement with the raw means in Table~\ref{tab:matched-raw} shows that the
result is not created by a few sequences contributing more windows.

\begin{table}[H]
\centering
\small
\begin{tabular}{rrrr}
\toprule
Sequence & Proposal & Grid & Difference \\
\midrule
41 & 0.6397 & 0.5629 & 0.0768 \\
42 & 0.4202 & 0.3823 & 0.0379 \\
43 & 0.5003 & 0.4466 & 0.0538 \\
44 & 0.3895 & 0.2972 & 0.0923 \\
45 & 0.3827 & 0.3147 & 0.0680 \\
46 & 0.3798 & 0.3749 & 0.0049 \\
47 & 0.3180 & 0.3170 & 0.0010 \\
48 & 0.3975 & 0.3676 & 0.0299 \\
\bottomrule
\end{tabular}
\caption{Language-path core BA averaged over the three fixed seeds within each validation sequence.}
\label{tab:matched-sequences}
\end{table}

The effect is positive on all eight sequences, although it is small on sequences 46 and 47. The result
supports proposal-selected temporal token locations over this fixed equal-count grid in the registered setting.
It does not establish detector accuracy, global optimality, or performance on an untouched test split.

\section{Radar Information Controls Across Objectives}

The main paper reports radar-input interventions for aligned checkpoints. Here we apply the same
evaluator to checkpoints trained with a fixed within-task verbalizer permutation and with no language
loss. Strict zero replaces the complete radar tensor, scene-local shuffle maps each window to another
window from the same sequence, repeat current duplicates the current sweep across history, reverse
history reverses the ten-sweep order, and zero Doppler removes the Doppler channel. The endpoint is
direct-head core BA. Positive real-minus-control values mean that the intervention reduces performance.

\begin{table}[H]
\centering
\scriptsize
\setlength{\tabcolsep}{2.6pt}
\begin{tabular}{llrrrr}
\toprule
Obj. & Input control & S41 & S42 & S43 & Mean \\
\midrule
P & Strict-zero radar & 0.1221 & 0.1434 & 0.1589 & 0.1415 \\
P & Scene-local shuffle & 0.1601 & 0.1824 & 0.1930 & 0.1785 \\
P & Repeat current & 0.0597 & 0.0520 & 0.0760 & 0.0626 \\
P & Reverse history & 0.0088 & 0.0164 & 0.0516 & 0.0256 \\
P & Zero Doppler & $-0.0024$ & 0.0069 & 0.0118 & 0.0054 \\
\addlinespace
N & Strict-zero radar & 0.1238 & 0.1541 & 0.1394 & 0.1391 \\
N & Scene-local shuffle & 0.1664 & 0.1973 & 0.1800 & 0.1812 \\
N & Repeat current & 0.0580 & 0.0544 & 0.0379 & 0.0501 \\
N & Reverse history & 0.0135 & 0.0289 & 0.0272 & 0.0232 \\
N & Zero Doppler & 0.0000 & 0.0089 & $-0.0054$ & 0.0012 \\
\bottomrule
\end{tabular}
\caption{Seed-level direct-head real-minus-control core BA. P denotes fixed verbalizer permutation, and
N denotes no language loss.}
\label{tab:objective-controls-seeds}
\end{table}

\begin{table}[H]
\centering
\scriptsize
\setlength{\tabcolsep}{2.5pt}
\begin{tabular}{lrrrr}
\toprule
Control & P mean & P 95\% CI & N mean & N 95\% CI \\
\midrule
Strict zero & 0.1415 & [0.0788, 0.2127] & 0.1391 & [0.0787, 0.2095] \\
Scene shuffle & 0.1785 & [0.0865, 0.2774] & 0.1812 & [0.0928, 0.2814] \\
Repeat current & 0.0626 & [0.0260, 0.0991] & 0.0501 & [0.0138, 0.0885] \\
Reverse history & 0.0256 & [0.0050, 0.0574] & 0.0232 & [0.0087, 0.0350] \\
Zero Doppler & 0.0054 & [$-0.0058$, 0.0219] & 0.0012 & [$-0.0112$, 0.0147] \\
\bottomrule
\end{tabular}
\caption{Crossed seed--sequence intervals for Table~\ref{tab:objective-controls-seeds}.}
\label{tab:objective-controls-ci}
\end{table}

For completeness, Table~\ref{tab:objective-language-diagnostic} reports the corresponding language-path
outputs. They are diagnostics rather than attribution endpoints: the permuted objective changes answer
strings by construction, and the no-language condition does not train the language path.

\begin{table}[t]
\centering
\scriptsize
\setlength{\tabcolsep}{2.5pt}
\begin{tabular}{lrrrr}
\toprule
Control & P mean & P 95\% CI & N mean & N 95\% CI \\
\midrule
Strict zero & $-0.0175$ & [$-0.0463$, 0.0099] & 0.0017 & [$-0.0110$, 0.0106] \\
Scene shuffle & $-0.0224$ & [$-0.0804$, 0.0199] & 0.0064 & [$-0.0007$, 0.0177] \\
Repeat current & $-0.0093$ & [$-0.0234$, 0.0041] & 0.0010 & [$-0.0029$, 0.0060] \\
Reverse history & $-0.0072$ & [$-0.0155$, 0.0014] & $-0.0004$ & [$-0.0024$, 0.0015] \\
Zero Doppler & 0.0021 & [$-0.0054$, 0.0118] & $-0.0009$ & [$-0.0032$, 0.0010] \\
\bottomrule
\end{tabular}
\caption{Secondary language-path real-minus-control BA with crossed intervals. P denotes fixed verbalizer
permutation, and N denotes no language loss. None of these diagnostic intervals excludes zero.}
\label{tab:objective-language-diagnostic}
\end{table}

Strict zero, scene-local shuffle, repeat current, and reverse history are positive for all three seeds
under both objectives, and their crossed intervals exclude zero. The direct predictor therefore uses
scene-specific radar content and multi-sweep temporal structure even when answer strings are permuted or
language loss is removed. Zero-Doppler effects are mixed and both intervals cross zero. These controls
establish objective-robust sensor dependence of the direct path, not a gain caused by aligned language
supervision.

\section{Robustness to Additional Fixed Verbalizers}

The primary audit in the main paper uses one frozen within-task verbalizer permutation. We trained three
additional task-local derangements, denoted P1--P3, from paired initial states for seeds 41--43. Each cell
uses the same Qwen2.5-3B checkpoint family, 1.2M projector cap, eight epochs, 8,192 samples per epoch,
and 4,208-window validation manifest. The mappings were frozen before training, and the completion
receipt records their SHA-256 hashes.

\begin{table}[H]
\centering
\small
\begin{tabular}{lrrrr}
\toprule
Fixed permutation & S41 & S42 & S43 & Mean \\
\midrule
P1 & 0.4871 & 0.5089 & 0.4861 & 0.4940 \\
P2 & 0.5124 & 0.5128 & 0.5114 & 0.5122 \\
P3 & 0.4836 & 0.5135 & 0.5037 & 0.5003 \\
\midrule
Within-seed range & 0.0288 & 0.0047 & 0.0253 & 0.0196 \\
\bottomrule
\end{tabular}
\caption{Direct-head core BA for three additional fixed verbalizer permutations.}
\label{tab:permutation-raw}
\end{table}

\begin{table}[H]
\centering
\small
\begin{tabular}{lrc}
\toprule
Pairwise contrast & Mean $\Delta$ & Sequence-cluster 95\% CI \\
\midrule
P1 $-$ P2 & $-0.0158$ & [$-0.0337$, 0.0019] \\
P1 $-$ P3 & $-0.0030$ & [$-0.0194$, 0.0141] \\
P2 $-$ P3 & 0.0129 & [0.0063, 0.0207] \\
\bottomrule
\end{tabular}
\caption{Pairwise mapping sensitivity conditional on the three fixed training seeds.}
\label{tab:permutation-pairwise}
\end{table}

The mean range across mappings is 0.0196 BA. The P2-minus-P3 interval excludes zero, whereas the other
pairwise intervals cross zero. Because these are three fixed derangements rather than
a random sample from a verbalizer population, and there are only three seeds, confirmatory seed-level
inference is ineligible. The result demonstrates descriptive mapping sensitivity. It does not overturn
the main paper's bounded aligned-versus-no-language conclusion.

\newpage
\section{Completion and Claim Boundary}

The terminal queue records $6/6$ verified matched-token cells, $6/6$ verified objective-control cells,
and $9/9$ verified additional-permutation cells, all on validation with no new test access. The receipt
binds the frozen protocols, ordered validation-window identity hash, source-tree hash, per-cell artifacts,
aggregates, and completion markers. Successful cells were not rerun during recovery. No result in this
supplement supports official nuScenes performance, K-Radar sequences 49--58 performance, detector AP,
unrestricted language generation, or globally optimal tokenization.

\end{document}